\documentclass[letterpaper, 10 pt, conference]{ieeeconf}  

\IEEEoverridecommandlockouts                              

\usepackage{amssymb}
\usepackage{amsmath}
\usepackage{biblatex}
\usepackage{graphicx}
\graphicspath{ {./figures/} }
\usepackage{float}
\usepackage{multirow}
\usepackage{fixltx2e}
\usepackage[usenames,dvipsnames]{color}
\usepackage{dblfloatfix}

\usepackage{balance}
\newcommand{\specialcell}[2][c]{%
  \begin{tabular}[#1]{@{}c@{}}#2\end{tabular}}

\title{\LARGE \bf
Body Gesture Recognition to Control a Social Robot
}

\author{Javier Laplaza, Joan Jaume Oliver, Ramón Romero, Alberto Sanfeliu and Anaís Garrell
\thanks{All authors work in the Institut de Robòtica i Informàtica Industrial de Barcelona (IRI), Catalonia, Spain
        {\tt\small javier.laplaza@upc.edu, anais.garrell@upc.edu}}%
}

\begin{document}

\maketitle
\thispagestyle{empty}
\pagestyle{empty}

\begin{abstract}
In this work, we propose a gesture based language to allow humans to interact with robots using their body in a natural way. We have created a new gesture detection model using neural networks and a custom dataset of humans performing a set of body gestures to train our network. Furthermore, we compare body gesture communication with other communication channels to acknowledge the importance of adding this knowledge to robots. The presented approach is extensively validated in diverse simulations and
real-life  experiments with non-trained volunteers. This  attains remarkable results and shows
that it is a valuable framework for social robotics applications,
such as human robot collaboration or human-robot interaction.
\end{abstract}

\section{Introduction}

Finding natural and efficient communication channels is essential in Human-Robot Interaction (HRI). If we take a look at the way humans communicate with each other, we see that about 70\% of the communication is non-verbal communication \cite{key_concepts_communication}. Moreover, when humans want to communicate with other agents with whom they do not share a common spoken language --foreigners, babys or  animals-- most of the communication is non-verbal~\cite{argyle1972non,hinde1972non}.

When it comes to communication with robots, it is possible to establish a set of gestures to communicate certain ideas in a similar way that gesture language works between humans. But, similarly to gesture language, this approach requires that both agents know which gestures compose the language and what meaning has each gesture. Furthermore, previous attempts of creating such body language with robots are generally designed by people used to work with robots. We argue that such kind of languages cannot be expected to find success when robots face humans with no previous experience  with robots.

This inspired us to seek a natural way for humans to communicate with robots using gestures. Our goal is to create a natural gesture dictionary and explore how humans rate gesture based communication versus other communication channels.

\begin{figure}
    \centering
    \includegraphics[width=\linewidth]{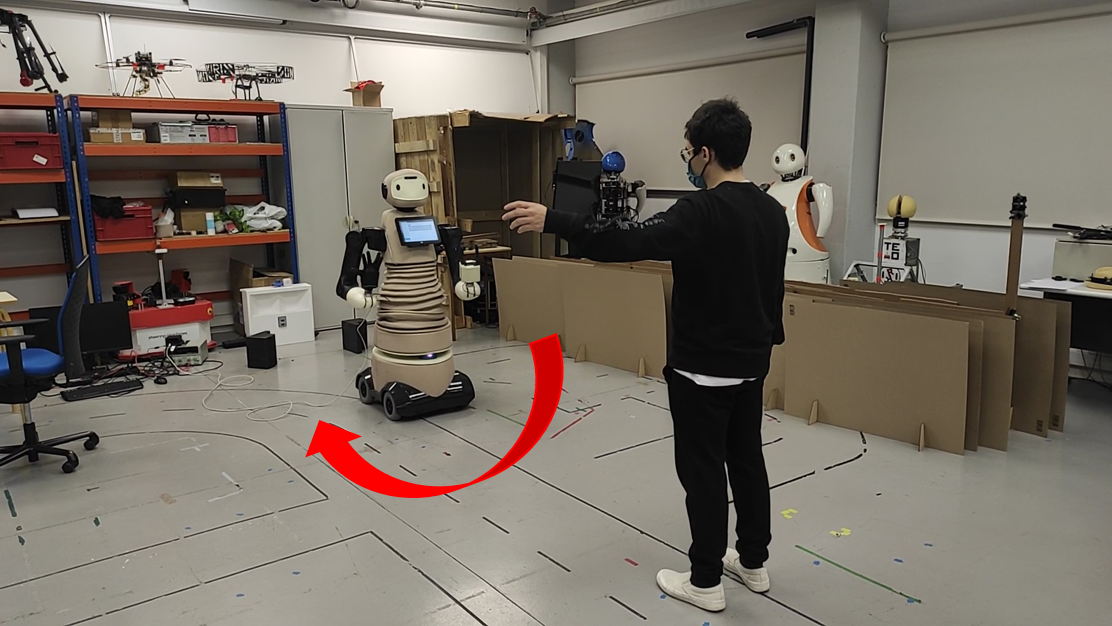}
    \caption{By raising the left arm, the human is able to tell the IVO robot to turn to the left side.}
    \label{fig:ivo}
\end{figure}

To achieve this, we first collected a dataset using human volunteers making a predefined set of communication gestures. Then, we create a gesture detection model to allow the robot identify human gestures. Finally, we wanted  to study how convenient was using gestures for communication. To do so, we carried out several experiments using non-trained volunteers and the IVO robot \cite{ivo}, see Fig.\ref{fig:ivo}, in order to compare gesture based communication to other communication channels.

The remainder of our study unfolds in the following manner. Section II introduces the related work in 
human-robot gesture recognition. Section III provides an overview of the contributions we describe in this article, presenting the model architecture. Section IV describes the created dataset the authors used to train and test the system.  In Section V, we present the setting of the  experiments and our evaluation methodology, which is subsequently employed in the results. Finally, conclusions are given in sections VI. 

\section{Related work}

Most of the literature related to gesture detection focuses on hand gesture detection. Great examples of this are \cite{how2sign}, \cite{hand_gesture_cnn} and \cite{hand_gesture_hri}.
Touchless interaction methods that will not use sound or speech for the communication need to somehow sense the human body's position. 

Wearable sensors are widely used in gesture recognition. For instance, incorporating IMUs in the wrist or some combination of them in the arms to track the body motion that will be
used to determine the robot commands, like in \cite{gromov2019proximity}, 
\cite{muhammad2020applications} and \cite{neto2019gesture}. Moreover, users can use data-gloves that track the hand and its finger's position,   \cite{gregory2019enabling} and \cite{pan2019sensor}.

Nevertheless, the fact of using specific wearables does not guarantee the touchless methods any advantages. Using gadgets on the human body takes back some of the touchless technologies strong points and also induces to a less natural communication.

This explains the snowball in research on communication with the bare human body, dressed as one would dress to interact with another human, without adding extra instruments or devices on the user.

In this particular work, we are interested in hand gestures and body gestures with the use of a camera, it  provides a richer language to express more complex concepts than full body gestures, and that might be the reason why not so many works focus on full body gesture communication with robots.

Some work on full body gesture communication focus on identifying where the human is pointing at to identify a specific object or region, one example of this being \cite{RAHEJA201814}. Other works focus on identifying emotions expressed by the body posture, such as \cite{gestures_emotions}.  Moreover, Some gestures are very restricted to specific tasks, in \cite{kwan2020gesture} the approach is to detect whether a human is willing to collaborate or not during a hand-over operation according to his gestural expression.

The work proposed in \cite{uav_gestures} is similar to our approach, but they use the relative position of hands and faces to define poses. More importantly, they use UAV as the platform to interact, which has different dynamics than ground robots and also a different perspective of humans when using a camera.

Another similar work is presented in \cite{miburi_dataset}, focusing in optical flow techniques used in order to extract features of the human body.

In \cite{proposed_gestures} a set of gestures are proposed for HRI operations. While very similar to our approach, in our work we focus on allowing very general gestures instead of defining them beforehand. Also, we study how human volunteers rate the interaction with the robot using gestures.

\section{Model Architecture}
In order to properly identify different body gestures, we have been working on a model that consists on a linear classifier with some hidden layers that are using different body positions as input paramaters while get gestures probability as outputs.

Although we are feeding our model with different body positions information, our inputs are videos with four dimensions $(x, y, z, t)$. In the work presented in this paper temporal component has been omitted and only the last frame of each video is used as input. This means that we only take into account those gestures that are static.

From this frame, we used MediaPipe Pose machine learning solution \cite{mediapipe} in order to extract the  full body coordinates \textit{(from head to feet)}.
This ML model predicts the location of 33 pose landmarks each following the $(x, y, z, visibility)$ structure that will be used as our Classifier input.


Once we pre-process the video to obtain our $[33, 4]$ matrix input containing all body positions coordinates, we can set up our classifier.

Our model uses four \textit{Linear} $y = xA^T + b$ data transformations fully connected layers of size $256, 128, 64$ and $8$ combined with ${ReLU}(x) = (x)^+ = \max(0, x)$ activation functions.

The outputs from our classifier consists in a one dimensional vector with length equals to the number of gestures that could be predicted, being each vector element the model score for each specific gesture.

From this vector, a ${Softmax}$ layer is applied in order to normalize the model scores and the maximum gesture probability is then considered to be the output gesture.

\begin{figure}[H]
    \includegraphics[width=\linewidth]{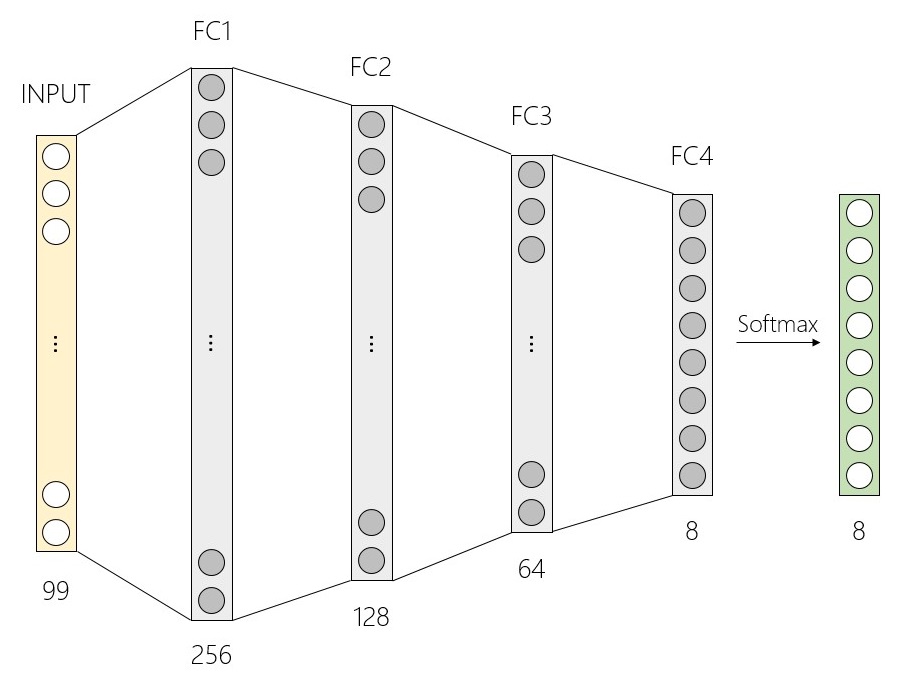}
    \caption{Schematic view of the neural network used for the gesture detection.}
    \label{fig:model}
\end{figure}

As for training and test strategies, and due to the low number of gestures and data available, instead of using the common $70\% - 30 \%$ strategy, the \textit{one out} method was used. More precisely, from the $8$ persons available all where used for train except for the $2$ one that was exclusively used for test purposes.

On one hand, it is interesting to say that the loss function used for training is the Cross-Entropy Loss, and on the other hand, the chosen optimization algorithm for this network is the Adaptive Moment Estimation (${Adam}$) with a learning rate $\alpha = 0.01$ and $\beta_1 = 0.9, \beta_2 = 0.999, \epsilon = 10^{-8}$ hyperparameters.

\section{Dataset}

Since we couldn't find a suitable dataset for our goal, we decided to create a dataset in our laboratory.

The main feature of our gesture based communication dictionary is naturalness. We want that everyone can communicate with the robot, not only people who is already familiar with robots. 

\begin{figure*}[h]
    \centering
    \includegraphics[scale=0.5]{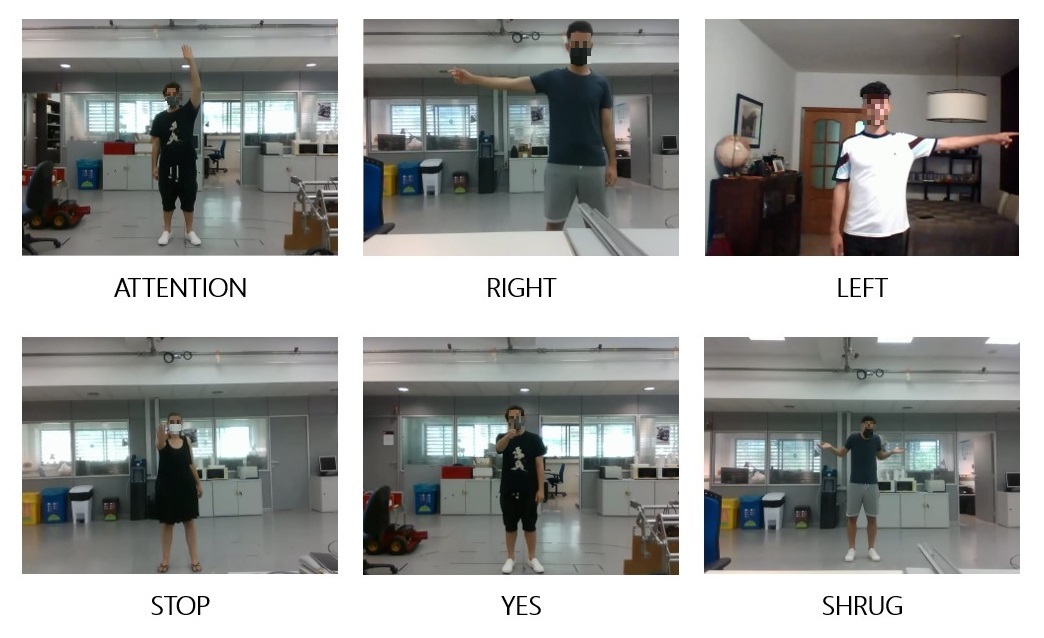}
    \caption{Some samples of static gestures recorded in the dataset.}
    \label{fig:gestures}
\end{figure*}

We divide the defined gestures in two groups: static and dynamic gestures (see Fig. \ref{fig:gestures}):

\begin{itemize}
    \item \textbf{Static gestures}
    \begin{itemize}
        \item \textbf{Attention}: Catch the robot's attention to give him an order.
        \item \textbf{Right}: Order the robot to turn right.
        \item \textbf{Left}: Order the robot to turn left.
        \item \textbf{Stop}: Order the robot to stop its trajectory.
        \item \textbf{Yes}: Approve a robot's information.
        \item \textbf{Shrug}: Inform the robot that you don't understand his information.
        \item \textbf{Random}: Random gesture, not necessarily a communication gesture.
        \item \textbf{Static}: Human is standing still.
    \end{itemize}
    \item \textbf{Dynamic gestures}
    \begin{itemize}
        \item \textbf{Greeting}: Greet the robot.
        \item \textbf{Continue}: Order the robot to continue its path after telling him to stop.
        \item \textbf{Turn-back}: Order the robot to turn 180 degrees.
        \item \textbf{No}: Deny a robot's information.
        \item \textbf{Slowdown}: Order the robot to reduce its speed.
        \item \textbf{Come}: Order the robot to reach your position.
        \item \textbf{Back}: Order the robot to move back.
    \end{itemize}
\end{itemize}

As for data recording, each human volunteer was recorded using an RGB camera. When human volunteers where asked to make a gesture they were provided with a vague explanation of the gesture intention. This was done to collect data that felt most natural to each volunteer. There was no restriction on which arms should be moved in each gesture whatsoever. Thus, different volunteers could make the same gesture in a very different way, using one arm or the other, or even both of them.

Each gesture was repeated three times, first 1 meter away from the camera, then 4 meters away and finally 6 meters away.Each video contains information of only one gesture, and all the videos were recorded indoors.

Finally, we use MediaPipe \cite{mediapipe} to extract the 3D joints of the human in the video.

We also consider a wide range of users regarding age, gender, education level and culture. Taking this into account, we created our dataset thanks to 10 human volunteers, 7 men and 3 women. Each volunteer was asked to perform 

\section{Experiments}
\subsection{Model results}
\begin{figure}
    \includegraphics[width=\linewidth]{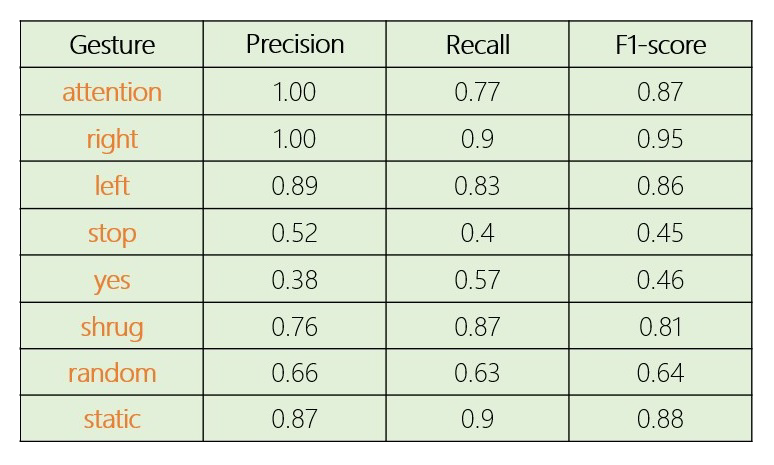}
    \caption{Precision, recall and F1-score for every model gesture class.}
    \label{fig:modelprecision}
\end{figure}

We train the model until we see that it starts overfitting in the test dataset. Then, we stop the training and check the results on this test dataset. These results can be seen in Table \ref{fig:modelprecision}.

The results show that the model accuracy for certain gestures is remarkable. Specifically, \textit{attention, right, left, shrug and static} gestures rate an F1-score above 0.8. This result was expected, since all those gestures are very different from other gestures in the dataset.

On the other hand, there are two gestures that the model particularly struggles classify: \textit{stop and yes}. Again, this result was expected, since both gestures present an overall body pose very similar, only the hand position differs.

Finally, the \textit{random} gesture has a F1-score of 0.64, which is understandable for a class that include very different body poses.

\begin{figure}
    \includegraphics[width=\linewidth]{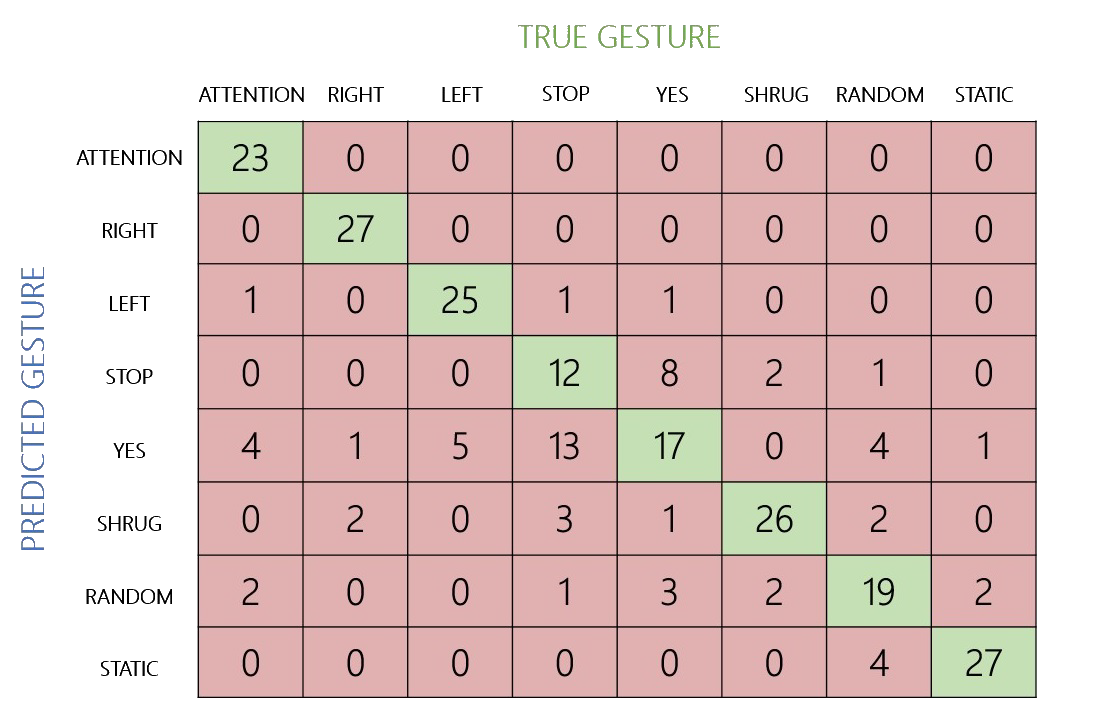}
    \caption{Model confussion matrix.}
    \label{fig:modelmatrix}
\end{figure}

In order to further study how do our defined gestures interfere with each other, we create a confusion matrix (see Fig. \ref{fig:modelmatrix}). From this matrix, we confirm that our model isn't able to classify gestures \textit{yes and stop}. Possible ways to tackle this can be increasing model complexity or using a more complete Mediapipe model that incorporates finger and face information.

\begin{figure*}[t]
    \centering
    \includegraphics[width=\textwidth]{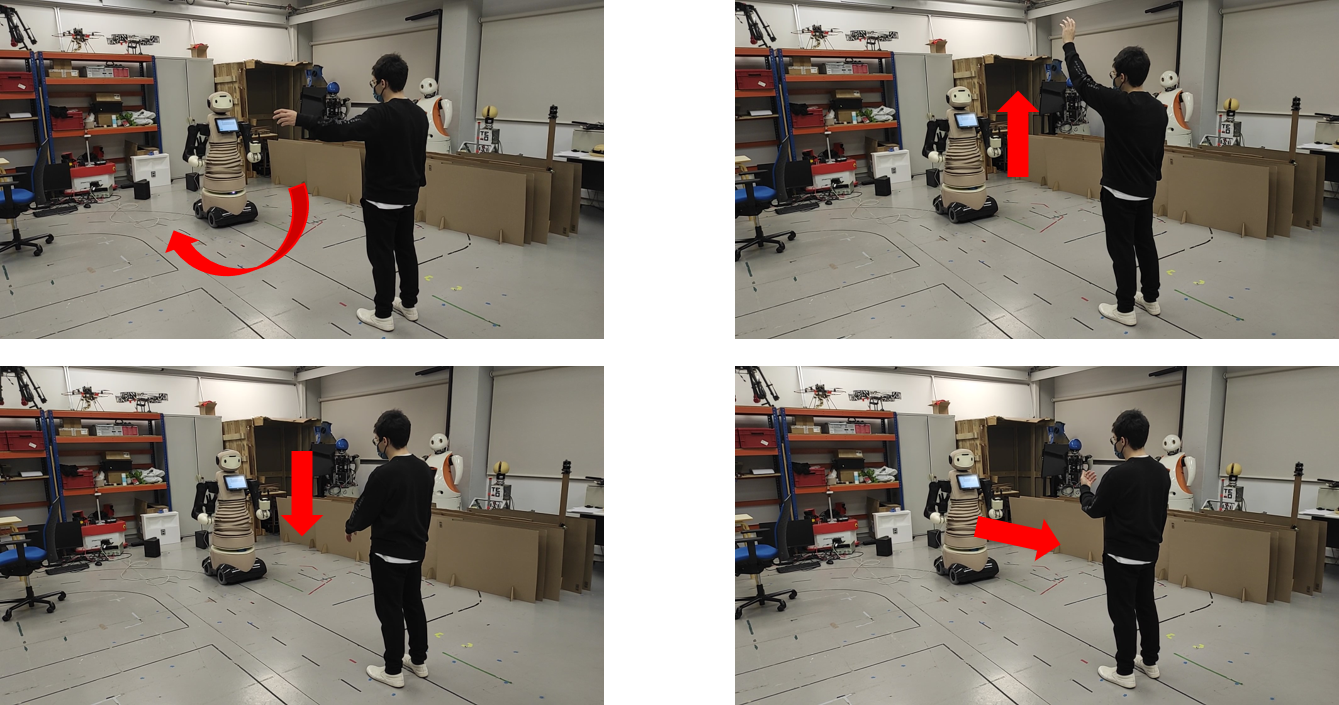}
    \caption{Samples of gestures used by the user: \textit{turn left} (top left), \textit{look upwards} (top right), \textit{look downwards} (bottom left) and \textit{approach} (bottom right).}
    \label{fig:user_samples}
\end{figure*}

\subsection{User Study}

The results presented in the previous section demonstrate that the robot is able to detect and recognize human natural gesture. A user study was also conducted to determine whether the body gesture recognition to control our robot enhances the usability and the comfort of the robot from the point of view of the human. We compared our method with the use of a remote controller. 

The hypothesis we endeavored to test was as follows: ``Participants will  feel more comfortable and will perceive  difference between the use of body gesture recognition and the use of a remote controller.''

We asked humans to communicate different orders to the robot, specifically:

\begin{itemize}
    \item Order the robot to move closer.
    \item Order the robot to move away.
    \item Order the robot to turn to the right.
    \item Order the robot to turn to the left.
    \item Order the robot to look up.
    \item Order the robot to look down.
\end{itemize}

In the first experiment, the human had to use the natural gestures to express to the robot what action must perform. We conducted these experiments in a Wizard-of-Oz way, since using the gesture detector may lead to missing some of the gestures, and it can cause a negative impact to the user perception of gestures as a channel of communication.

Then, we repeated the same experiment but this time we gave a remote controller to the human, thus  that he/she could  tele-operate the robot after some instruction.

In each case, we asked the volunteers to make all gestures/commands in a random order. We also chose randomly between the gesture communication and the controller as the first experiment for each volunteer in order to avoid possible biases.
Refer to Fig.\ref{fig:user_samples} to seen  some samples of the gesture communication.

\begin{figure*}[t]
    \centering
    \includegraphics[width=0.98\textwidth]{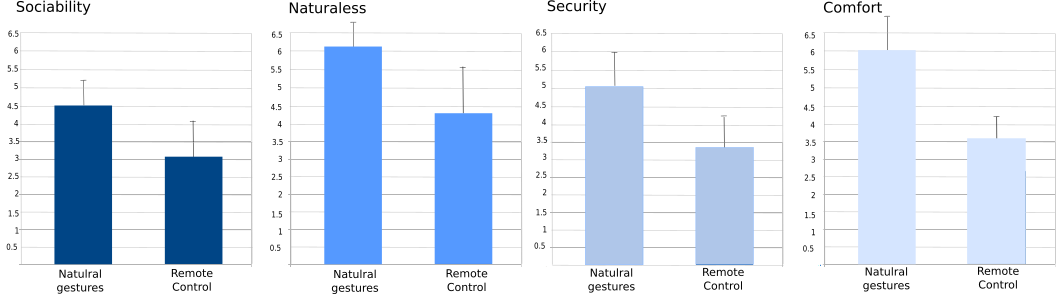}
    \caption{Evaluation from 1 (low)
to 7 (high) of the main aspects related to the robot behavior in handover task.}
    \label{fig:hri}
\end{figure*}


For the experiments, we selected 15 people (8 men, 7 women) on the University Campus. Participants ranged in age from 19 to 50 years (M=29.5, SD=9.2), and represented a variety of University
majors and occupations including computer science, mathematics, biology, finance and chemistry. For each individual selected, we randomly activated one of the two  behaviors to deliver an object to the volunteer, it is, the activation of the prediction module or the not use of it.
It should be mentioned that none of the participants had previous experience working or interacting with robots.

Participants were asked to complete a variety of surveys. Our independent variables considered whether participant make use of our gesture recongnition or the remote control. The main dependent
variables involved participants' perceptions of the \textbf{sociability}, \textbf{naturalness}, \textbf{security} and \textbf{comfort}  characteristics.
Each of these fields, was evaluated by every participant using a questionnaire to fill out after the
experiment, see Table \ref{tab:survey_questions}, based on~\cite{kirby2010social}.

Participants were asked to answer a questionnaire, following their encounter with the robot in each mode of behavior. To analyze their responses, we grouped the
survey questions into four scales: the first measured sociability robot behavior, while the second  naturalness, and third and fourth  evaluated the comfort. Both scales surpassed the commonly used 0.7 level of reliability (Cronbach's alpha).

Each scale response was computed by averaging the results of the survey questions comprising the scale. ANOVAs were run on each scale to highlight differences between the three robot behaviors.

Below, we provide the results of comparing the two different methods. To analyze the source of the difference, four scores were examined: ``sociability'', ``naturalness'', ``security'' and ``comfort'', plotted in Fig.~\ref{fig:hri}. For all four aspects, the evaluation score plotted in Fig.~\ref{fig:hri}, pairwise comparison with Bonferroni demonstrate there were   difference between the two kind of behavior approaches, $p<0.05$.

Therefore, after analyzing these four components,  we may conclude that if the robot is capable of understand people's body gesture  the acceptability of the robots increases, and participants perceived the robot as a social entity.

\begin{table}[]
    \centering
    \begin{tabular}{|c|}
    
    \hline
    \textbf{Question} \\
    \hline
    \hline
         \specialcell{How \textbf{positive/negative} did you feel \\ about your interaction with the robot?} \\
         How \textbf{human-like} did the robot behave? \\
         How \textbf{social} was the robot's behavior? \\
         \hline
         \hline
         \specialcell{How \textbf{hopeful/despairing} did you feel \\ about your interaction with the robot?} \\
         How \textbf{natural} was the robot's behavior? \\
         \hline
         \hline
         \specialcell{How \textbf{excited/calm} did you feel \\ about your interaction with the robot?} \\
         How \textbf{safe} did you feel around the robot? \\
         \hline
         \hline
         \specialcell{How much \textbf{control} did you feel \\ about your interaction with the robot?} \\
         \specialcell{How \textbf{agitated/comfortable} did you feel \\ about your interaction with the robot?} \\
         How \textbf{comfortable} did you feel near the robot? \\
         \hline

    \end{tabular}
    \caption{Each block of questions correspond to the quality being measured. From top to bottom: sociability, naturalness, security and comfort.}
    \label{tab:survey_questions}
\end{table}

\section{Conclusions}

We created a dataset of natural gestures to communicate with a robot. We also developed a new neural network architecture able to classify the static gestures of the dataset. We reach high classification accuracy (>80\%)) in most of the gestures, except for gestures that look very similar. Our future work will try to add also the dynamic gestures to the model and decouple gestures that can be easily confused.

The findings presented in the previous section reinforce the notion that the robot’s ability to understand human body gestures is an important skill to master in order to achieve natural interaction with people. Overall, people  interacting and communicating with the robot using natural gestures enhances the engagement between the robot and the human.

The experiments we conducted yielded conclusive results. We found that people felt their interaction with
the robot was more natural when the robot communicated through gestures. Detailed analysis showed that these capacities improved the human’s perception of the robot’s security and sociability. We also found that the amount of speech and comments made by the robot seems to be appropriate for this type of scenario.  Finally, volunteers perceived our robot more sociable and closer when it recognized and understood  their gestures, and the interactions were longer. 


\balance
\printbibliography

@book{key_concepts_communication,
  title     = {Key concepts in communication and cultural studies},
  author    = {Tim O'Sullivan and John Hartley and Danny Saunders and Martin Montgomery and John Fiske},
  isbn      = {0415061733},
  year      = {1994},
  publisher = {Routledge}
}

@article{how2sign,
  author    = {Amanda Cardoso Duarte and
               Samuel Albanie and
               Xavier Gir{\'{o}}{-}i{-}Nieto and
               G{\"{u}}l Varol},
  title     = {Sign Language Video Retrieval with Free-Form Textual Queries},
  journal   = {CoRR},
  volume    = {abs/2201.02495},
  year      = {2022},
  url       = {https://arxiv.org/abs/2201.02495},
  eprinttype = {arXiv},
  eprint    = {2201.02495},
  bibsource = {dblp computer science bibliography, https://dblp.org}
}

@InProceedings{hand_gesture_cnn,
author="Alashhab, Samer
and Gallego, Antonio-Javier
and Lozano, Miguel {\'A}ngel",
editor="De La Prieta, Fernando
and Omatu, Sigeru
and Fern{\'a}ndez-Caballero, Antonio",
title="Hand Gesture Detection with Convolutional Neural Networks",
booktitle="Distributed Computing and Artificial Intelligence, 15th International Conference",
year="2019",
publisher="Springer International Publishing",
address="Cham",
pages="45--52",
isbn="978-3-319-94649-8"
}

@article{hand_gesture_hri,
title = {A hand gesture recognition technique for human–computer interaction},
journal = {Journal of Visual Communication and Image Representation},
volume = {28},
pages = {97-104},
year = {2015},
issn = {1047-3203},
doi = {https://doi.org/10.1016/j.jvcir.2015.01.015},
url = {https://www.sciencedirect.com/science/article/pii/S104732031500022X},
author = {Nurettin Çağrı Kılıboz and Uğur Güdükbay}
}

@unknown{mediapipe,
author = {Lugaresi, Camillo and Tang, Jiuqiang and Nash, Hadon and McClanahan, Chris and Uboweja, Esha and Hays, Michael and Zhang, Fan and Chang, Chuo-Ling and Yong, Ming and Lee, Juhyun and Chang, Wan-Teh and Hua, Wei and Georg, Manfred and Grundmann, Matthias},
year = {2019},
month = {06},
pages = {},
title = {MediaPipe: A Framework for Building Perception Pipelines}
}

@ARTICLE{gestures_emotions,
  author={Noroozi, Fatemeh and Corneanu, Ciprian Adrian and Kamińska, Dorota and Sapiński, Tomasz and Escalera, Sergio and Anbarjafari, Gholamreza},
  journal={IEEE Transactions on Affective Computing}, 
  title={Survey on Emotional Body Gesture Recognition}, 
  year={2021},
  volume={12},
  number={2},
  pages={505-523},
  doi={10.1109/TAFFC.2018.2874986}
  }

@article{RAHEJA201814,
title = {3D gesture based real-time object selection and recognition},
journal = {Pattern Recognition Letters},
volume = {115},
pages = {14-19},
year = {2018},
note = {Multimodal Fusion for Pattern Recognition},
issn = {0167-8655},
doi = {https://doi.org/10.1016/j.patrec.2017.09.034},
url = {https://www.sciencedirect.com/science/article/pii/S0167865517303549},
author = {Jagdish Lal Raheja and Mona Chandra and Ankit Chaudhary}
}

@inproceedings{ivo,
author = {Laplaza, Javier and Rodr\'{\i}guez, Nicol\'{a}s and Dom\'{\i}nguez-Vidal, J. E. and Herrero, Fernando and Hern\'{a}ndez, Sergi and L\'{o}pez, Alejandro and Sanfeliu, Alberto and Garrell, Ana\'{\i}s},
title = {IVO Robot: A New Social Robot for Human-Robot Collaboration},
year = {2022},
publisher = {IEEE Press},
booktitle = {Proceedings of the 2022 ACM/IEEE International Conference on Human-Robot Interaction},
pages = {860–864},
numpages = {5},
location = {Sapporo, Hokkaido, Japan},
series = {HRI '22}
}

@INPROCEEDINGS{uav_gestures,
  author={MohaimenianPour, Sepehr and Vaughan, Richard},
  booktitle={2018 IEEE/RSJ International Conference on Intelligent Robots and Systems (IROS)}, 
  title={Hands and Faces, Fast: Mono-Camera User Detection Robust Enough to Directly Control a UAV in Flight}, 
  year={2018},
  volume={},
  number={},
  pages={5224-5231},
  doi={10.1109/IROS.2018.8593709}}

@article{miburi_dataset,
  author    = {Jen{-}Yen Chang and
               Antonio Tejero{-}de{-}Pablos and
               Tatsuya Harada},
  title     = {Improved Optical Flow for Gesture-based Human-robot Interaction},
  journal   = {CoRR},
  volume    = {abs/1905.08685},
  year      = {2019},
  url       = {http://arxiv.org/abs/1905.08685},
  eprinttype = {arXiv},
  eprint    = {1905.08685},
  bibsource = {dblp computer science bibliography, https://dblp.org}
}

@article{proposed_gestures,
  author    = {Jia Chuan A. Tan and
               Wesley P. Chan and
               Nicole L. Robinson and
               Elizabeth A. Croft and
               Dana Kulic},
  title     = {A Proposed Set of Communicative Gestures for Human Robot Interaction
               and an {RGB} Image-based Gesture Recognizer Implemented in {ROS}},
  journal   = {CoRR},
  volume    = {abs/2109.09908},
  year      = {2021},
  url       = {https://arxiv.org/abs/2109.09908},
  eprinttype = {arXiv},
  eprint    = {2109.09908},
  bibsource = {dblp computer science bibliography, https://dblp.org}
}

@misc{kwan2020gesture,
      title={Gesture Recognition for Initiating Human-to-Robot Handovers}, 
      author={Jun Kwan and Chinkye Tan and Akansel Cosgun},
      year={2020},
      eprint={2007.09945},
      archivePrefix={arXiv},
      primaryClass={cs.CV}
}

@book{hinde1972non,
  title={Non-verbal communication},
  author={Hinde, Robert A and Hinde, Robert Aubrey},
  year={1972},
  publisher={Cambridge University Press}
}

@article{argyle1972non,
  title={Non-verbal communication in human social interaction.},
  author={Argyle, Michael},
  year={1972},
  publisher={Cambridge U. Press}
}

@inproceedings{gromov2019proximity,
  title={Proximity human-robot interaction using pointing gestures and a wrist-mounted IMU},
  author={Gromov, Boris and Abbate, Gabriele and Gambardella, Luca M and Giusti, Alessandro},
  booktitle={2019 International Conference on Robotics and Automation (ICRA)},
  pages={8084--8091},
  year={2019},
  organization={IEEE}
}

@inproceedings{muhammad2020applications,
  title={Applications of Myo Armband Using EMG and IMU Signals},
  author={Muhammad, Uzair and Sipra, Khadija Amjad and Waqas, Muhammad and Tu, Shanshan},
  booktitle={2020 3rd International Conference on Mechatronics, Robotics and Automation (ICMRA)},
  pages={6--11},
  year={2020},
  organization={IEEE}
}

@article{gregory2019enabling,
  title={Enabling intuitive human-robot teaming using augmented reality and gesture control},
  author={Gregory, Jason M and Reardon, Christopher and Lee, Kevin and White, Geoffrey and Ng, Ki and Sims, Caitlyn},
  journal={arXiv preprint arXiv:1909.06415},
  year={2019}
}

@inproceedings{pan2019sensor,
  title={A Sensor Glove for the Interaction with a Nursing-Care Assistive Robot},
  author={Pan, Shimin and Lv, Honghao and Duan, Hong and Pang, Gaoyang and Yi, Kang and Yang, Geng},
  booktitle={2019 IEEE International Conference on Industrial Cyber Physical Systems (ICPS)},
  pages={405--410},
  year={2019},
  organization={IEEE}
}

@article{neto2019gesture,
  title={Gesture-based human-robot interaction for human assistance in manufacturing},
  author={Neto, Pedro and Sim{\~a}o, Miguel and Mendes, Nuno and Safeea, Mohammad},
  journal={The International Journal of Advanced Manufacturing Technology},
  volume={101},
  number={1},
  pages={119--135},
  year={2019},
  publisher={Springer}
}

\end{document}